\documentclass[10pt,twocolumn,letterpaper]{article}

\usepackage{iccv}
\usepackage{times}
\usepackage{epsfig}
\usepackage{graphicx}
\usepackage{amsmath}
\usepackage{amssymb}
\usepackage{stackengine}
\usepackage{textcomp}
\usepackage{multirow}
\usepackage[ruled,vlined]{algorithm2e}
\usepackage{comment}
\newcommand\Dtilde{\stackrel{\sim}{\smash{\text{D}}\rule{0pt}{1.1ex}}}

\usepackage[breaklinks=true,bookmarks=false]{hyperref}
\hypersetup{
    colorlinks=true,
    filecolor=magenta,      
    urlcolor=magenta,
}

\iccvfinalcopy 

\def\iccvPaperID4358Enter the ICCV Paper ID here
\def\httilde{\mbox{\tt\raisebox{-.5ex}{\symbol{126}}}}

\ificcvfinal\fi

\begin{document}


\title{Teacher-Student Adversarial Depth Hallucination to Improve Face Recognition}

\author{Hardik Uppal~~~~~~~~~Alireza Sepas-Moghaddam~~~~~~~~~Michael Greenspan~~~~~~~~~Ali Etemad\\
Queen's University, Canada}




\maketitle
\ificcvfinal\thispagestyle{empty}\fi

\begin{abstract}
We present the Teacher-Student Generative Adversarial Network (TS-GAN) to generate depth images from single RGB images in order to boost the performance of face recognition systems. For our method to generalize well across unseen datasets, we design two components in the architecture, a teacher and a student. The teacher, which itself consists of a generator and a discriminator, learns a latent mapping between input RGB and paired depth images in a supervised fashion. The student, which consists of two generators (one shared with the teacher) and a discriminator, learns from new RGB data with no available paired depth information, for improved generalization. The fully trained shared generator can then be used in runtime to hallucinate depth from RGB for downstream applications such as face recognition. We perform rigorous experiments to show the superiority of TS-GAN over other methods in generating synthetic depth images. Moreover, face recognition experiments demonstrate that our hallucinated depth along with the input RGB images boost performance across various architectures when compared to a single RGB modality by average values of +1.2\%, +2.6\%, and +2.6\% for IIIT-D, EURECOM, and LFW datasets respectively. We make our implementation public at: \href{https://github.com/hardik-uppal/teacher-student-gan.git}{https://github.com/hardik-uppal/teacher-student-gan.git}.

\end{abstract}

\section{Introduction}
Facial recognition is an active research area, which has recently witnessed considerable progress thanks primarily to the effectiveness of deep neural networks such as AlexNet~\cite{krizhevsky2017imagenet}, VGG~\cite{simonyan2014very}, FaceNet~\cite{schroff2015facenet}, ResNet~\cite{he2016deep} and others. 
RGB-based face recognition methods tend to be generally sensitive to facial and environmental variations like illumination, occlusions, and poses~\cite{sengupta2016frontal,zhang2018improving,adini1997face,mehdipour2016comprehensive}. 
Utilizing the depth information, acquired with an RGB-D sensor such as the Microsoft Kinect or Intel Realsense, alongside RGB allows models to learn more robust face representations. This is because depth provides complementary geometric information about the intrinsic shape of the face, further boosting recognition performance. Additionally, RGB-D facial recognition methods are known to be less sensitive than pure RGB approaches to pose and illumination variations~\cite{uppal2020attention,Chowdhury2016rgbd,Hayat2016rgbd,uppal2021depth}. Despite these advantages, while RGB sensors are ubiquitous, depth sensors have been less prevalent, resulting in an over-reliance on RGB alone. To tackle this, we present a method that uses available paired RGB-D training data to learn to \textit{hallucinate} (i.e. generate synthetic) depth images, even for datasets for which corresponding ground-truth depth information is absent.

\begin{figure}
    \centering
    \includegraphics[width=\columnwidth]{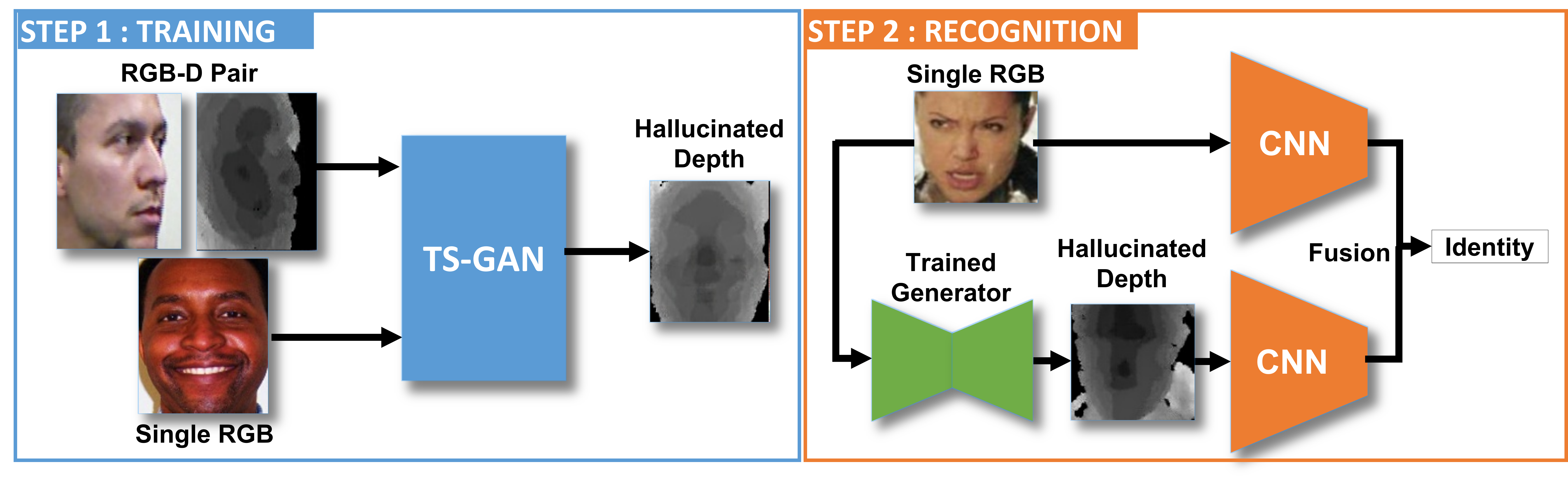}
    \caption{The proposed framework for our method. The first step (blue) trains the generator for synthesizing depth from RGB images, while the second step (orange) tests the efficacy of the synthesized depth images by using it in face recognition pipelines.}
    
    \label{fig:banner}
\end{figure}

Generative Adversarial Networks (GANs)~\cite{goodfellow2014generative} and its variants (e.g., cGan~\cite{mirza2014conditional}, pix2pix~\cite{isola2017image}, CycleGan~\cite{zhu2017unpaired}, StackGAN~\cite{zhang2017stackgan}, StyleGAN~\cite{karras2019style}, etc.) have proven to be viable solutions for data synthesis in many application domains. In the context of facial images, GANs have been widely used to generate very high-quality RGB images when trained on large-scale datasets such as FFHQ~\cite{karras2019style,karras2020analyzing} and CelebA-HQ~\cite{karras2017progressive}. Nonetheless, only a limited number of past works have attempted to synthesize depth from corresponding RGB images using a conditional GAN~\cite{pini2018learning}, CycleGAN~\cite{kwak2020novel}, and a Fully Convolutional Network (FCN)~\cite{cui2018improving_newadd}. Although cGAN has achieved impressive results for depth synthesis using paired RGB-D sets~\cite{pini2018learning}, it does not easily generalize to new test examples for which paired samples are not available, especially when the images are from an entirely different dataset with drastically different poses, expressions, and occlusions. CycleGAN~\cite{zhu2017unpaired} attempts to overcome this shortcoming through unpaired training with the aim of generalizing well to new test examples. However, as stated in~\cite{zhu2017unpaired}, CycleGAN does not deal well with translating geometric shapes and features.

In this work, we propose a deep architecture using a novel Teacher-Student GAN (TS-GAN) to generate depth images from RGB images for which no corresponding depth information is available. Our end-to-end model consists of two components, a teacher and a student. The teacher consists of a fully convolutional encoder-decoder network as a generator along with a fully convolutional classification network as the discriminator. The generator takes RGB images as inputs and aims to output the corresponding depth images. In essence, our teacher aims to learn an initial latent mapping between RGB and co-registered depth images. The student consists of two generators in the form of encoder-decoders, one of which is shared with the teacher, along with a fully convolutional discriminator. The student takes as its input an RGB image for which the corresponding depth image is not available and maps it onto the depth domain as guided by the teacher. The purpose here is for the student to further refine the strict mapping learned by the teacher and allow for better generalization through a less constrained training scheme. We demonstrate the high quality of our hallucinated depth images by comparing them to ground truth depth and several state-of-the-art depth generation alternatives. The performance of our approach for using the generated depth in facial recognition is then validated for two RGB-D datasets, IIIT-D RGB-D and EURECOM KinectFaceDb, across various facial recognition networks. The results show that the depth images generated using our approach enable a performance as good as, or in some cases surprisingly even better than using the ground-truth depth originally available in the dataset, and that it gives a significant boost to recognition accuracy as compared to a pure RGB facial recognition system. We also evaluate the performance of our approach for an in-the-wild RGB dataset, Labeled-Faces-in-Wild (LFW), where no depth information is originally available, and show that the addition of hallucinated depth by our proposed method can considerably boost the recognition results by +2.4\% with SE-ResNet-50 architecture.

Our contributions are summarized as follows. (\textbf{1}) A novel teacher-student adversarial architecture is proposed to generate realistic depth images from a single RGB image. Our method uses a student architecture to refine the strict latent mapping between RGB and D domains learned by the teacher to obtain a more generalizable and less constrained relationship. (\textbf{2}) Our assessments reveal that our method creates realistic synthetic depth images as compared to the original co-registered depth images (where available) and other techniques. We then utilize the synthetic depth for RGB-D facial recognition and show that multimodal solutions that utilize the depth images produced by our method perform as good as using the ground-truth depths. We also show that the facial recognition performance increases when utilizing our method to generate depth for an RGB-only dataset and subsequently combining the generated depth and original RGB images in a multimodal network. (3) We make our implementation publicly available\footnote{https://github.com/hardik-uppal/teacher-student-gan.git} to enable reproducibility and future comparisons.

\section{Related Work}

\subsection{Depth Generation from RGB Images}
\label{sec:depth_gen_review}
A number of methods have been proposed to estimate depth information from other modalities such as stereo vision~\cite{godard2019digging,godard2017unsupervised,badki2020bi3d} and multi-view images~\cite{wang2018mvdepthnet}. Here, given our goal in this paper, we only review methods that generate depth images from RGB data. 

The majority of existing work in this area relies on classical non-deep techniques. Sun~\etal \cite{sun2011depth} used images of different 2D face poses to create a 3D model. This was achieved by calculating the rotation and translation parameters with constrained independent component analysis and combining it with a prior 3D model for depth estimation of specific feature points. In a subsequent work~\cite{sun2012depth}, a nonlinear least-squares model was exploited to predict the depth of specific facial feature points, thereby inferring the 3D structure of the face. Both these methods used facial landmarks obtained by detectors for parameter initialization, making them highly dependent on landmark detection. Liu \etal \cite{liu2014discrete} modelled image regions as superpixels and used optimization for depth estimation. In this context, the continuous variables encoded the depth of the superpixel while the discrete variables represented their internal relationships. In a later work, Zhu \etal~\cite{zhuo2015indoor} exploited the global structure of the scene by constructing a hierarchical representation of local, mid-level, and large-scale layouts. They modeled the problem as conditional Markov random field with variables for each layer in the hierarchy. In~\cite{kong2016effective}, Kong \etal mapped a 3D dataset to 2D images by sampling points from the dense 3D data and combining them with RGB channel information. They then exploited face Delaunay triangulation to create a structure of facial feature points. The similarity of the triangles among the test images and the training set allowed them to estimate depth.

A few methods have attempted to synthesize depth using deep learning architectures. Cui \etal~\cite{cui2018improving_newadd} estimated depth from RGB using a multi-task approach consisting of face identification along with depth estimation. They also performed RGB-D recognition experiments to study the effectiveness of the estimated depth for the recognition task using an Inception-V2~\cite{ioffe2015batch} fusion network on the Lock3dFace and IIIT-D RGB-D datasets. Pini~\etal~\cite{pini2018learning} used a cGAN architecture for facial depth map estimation from monocular intensity images. Their method used co-registered intensity and depth images to train a generator in order to learn the relationship between RGB and depth images for face verification. Kwak ~\etal~\cite{kwak2020novel} proposed a solution based on CycleGAN~\cite{zhu2017unpaired} for generating depth and image segmentation maps. To estimate depth information, the characteristics of input RGB images were maintained with the help of the consistency loss of CycleGAN. This was aided through a multi-task approach by generating segmentation maps for those RGB images which would further help the network to fill in depth information where it was ambiguous or hidden by overlapping of features of the image.

\subsection{RGB-D Face Recognition}
Early RGB-D facial recognition methods were proposed based on classical (non-deep) methods. Goswami \etal~\cite{Goswamirgbd} fused visual saliency and entropy maps extracted from RGB and depth data. Histograms of oriented gradients were then used to extract features from image patches to then feed a classifier for identity recognition. Li \etal~\cite{li2016face} used 3D point-cloud data to obtain a pose-corrected frontal view using a discriminant color space transformation. The corrected texture and depth maps were sparse approximated using separate dictionaries that were learned during the training phase. Hayat \etal~\cite{Hayat2016rgbd} used a co-variance matrix representation on the Riemannian manifold to extract independent features from RGB and depth data, followed by an SVM classifier with score-level fusion to classify identities.

Recent methods have mainly focused on deep neural networks for RGB-D facial recognition. Chowdhury \etal~\cite{Chowdhury2016rgbd} used Auto-Encoders to learn a mapping function between RGB and depth. The mapping function was then used to reconstruct depth images from the corresponding RGB to be used for identification. Zhang \etal~\cite{Zhang2018afgr} tackled the problem of multi-modal recognition using deep learning, focusing on joint learning of the CNN embedding to fuse the common and complementary information offered by the RGB and depth together effectively.

In~\cite{LFFace}, RGB, disparity maps, and depth images were independently used to fine-tune separate VGG-Face~\cite{Parkhi15} models. The obtained embeddings were then fused to feed an SVM classifier for performing facial recognition. Jiang \etal\cite{Jiang_PAMI} proposed an attribute-aware loss function for CNN-based facial recognition which aimed to regularize the distribution of learned representations with respect to soft-biometric attributes such as gender, ethnicity, and age, thus boosting recognition results. Lin \etal\cite{lin2020rgb_newadd} proposed an RGB-D face identification method by introducing new loss functions, including associative and discriminative losses, which were then combined with softmax loss for training, showing boosted recognition results on the IIIT-D RGB-D dataset. Uppal \etal~\cite{uppal2021depth} proposed a two-level attention module to fuse RGB and depth modalities. The first attention layer selectively focused on the fused feature maps obtained by a convolutional feature extractor that were recurrently learned by an LSTM layer. The second attention layer then focused on the spatial features of those maps by applying attention weights using a convolution layer. In~\cite{uppal2021depth}, the authors proposed an attention-based method in which the features of depth images allowed the network to focus on regions of the face in the RGB images that contained prominent person-specific information.

\begin{figure*}[!ht]          
\centering
\includegraphics[width=0.95\textwidth]{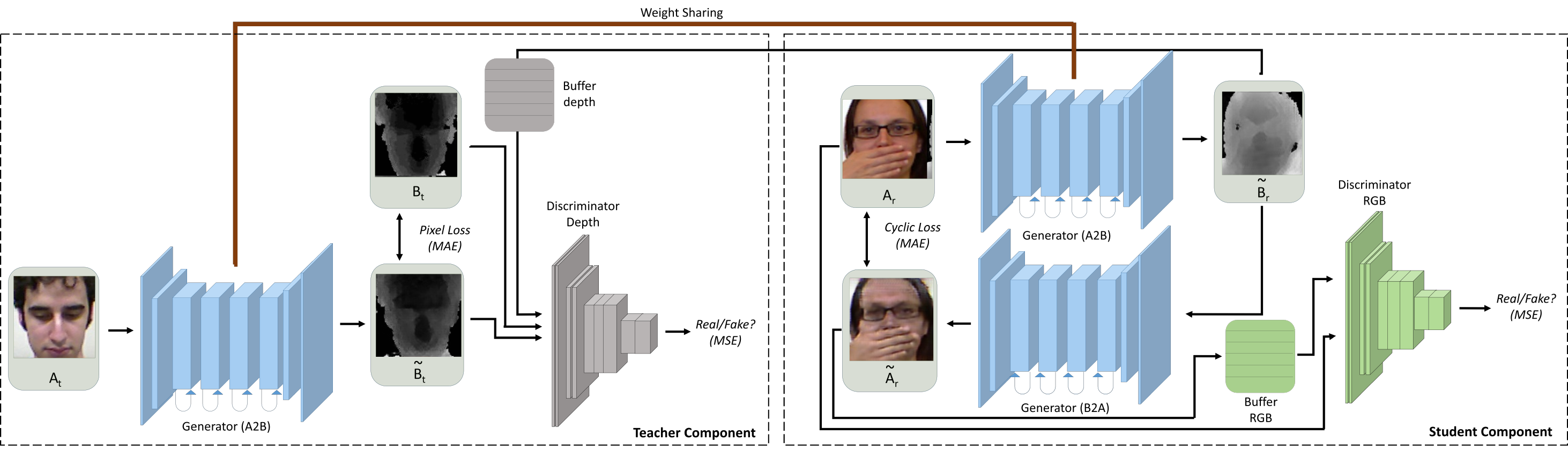}
\caption{The architecture details for our proposed teacher-student adversarial network are presented. $A_{t}$ and $B_t$ refer to the co-registered RGB and depth images respectively, and $\widetilde{B_t}$ refers to the generated depth in the teacher component. $A_r$ refers to the RGB image (when no corresponding depth is available), and $\widetilde{A_r}$ denotes the reconstructed RGB image. $\widetilde{B_r}$ refers to the hallucinated depth generated by our model for the particular RGB image.}
\label{fig:architecture}
\end{figure*}

\section{Method}
\label{sec:method}
\subsection{Problem Formulation}

We consider the problem of depth generation for a target dataset of RGB images $\{A_{r}\}_{i=1}^M$, whose distribution is $A_{r} \sim p_{target}(A_{r})$, and have no corresponding depth information. Let's assume we are provided with an RGB-D dataset which we refer to as the teacher dataset $\{A_{t}, B_{t}\}_{i=1}^N$ with distribution $A_{t}, B_{t} \sim p_{train}(A_{t},B_{t})$, with $A_{t}$ being an RGB image and $B_{t}$ being the co-registered depth image. Our goal is to learn from the teacher dataset a mapping generator function $G_{A2B}$ that can accurately generate an estimated depth image $\widetilde{B}_r$ for each target RGB image $A_{r}$. 

\subsection{Loss Formulation and Algorithm}
\label{sec:losses}

Our end-to-end architecture TS-GAN consists of a teacher component and a student component. The aim of the teacher, which itself consists of a generator and a discriminator, is to learn a latent mapping between $A_t$ and $B_t$. The student then refines the learned mapping for $A_r$ by further training the generator, with the aid of another generator-discriminator pair. Figure~\ref{fig:architecture} presents the TS-GAN architecture. For the teacher we create a mapping function, $G_{A2B}$:$A_{t}$$\,\to\,$$B_{t}$ along with a binary discriminator function $D_{depth}(.)$, which classifies whether the input is a real or fake (generated depth image). The loss $\mathcal{L}_{G_{A2B}}$ for the mapping function is then formulated as:
\begin{multline}
\small
\begin{aligned}
\hspace{-0.3 cm}
~\mathcal{L}_{G_{A2B}} = 
 \frac{1}{2} \mathbb{E}_{A_{t} \sim p_{train}(A_{t})}[ (D_{depth}(G_{A2B}(A_{t})) - 1)^2],
\end{aligned}
\label{eq:teach_net_gen}
\end{multline}
where $\mathbb{E}_{A_{t} \sim p_{train}(A_{t})}$ represents an RGB image sampled from $p_{train}(A_{t})$, the distribution of RGB images in the teacher dataset.

The loss $\mathcal{L}_{D_{depth}}$ for the depth discriminator, whose goal is to differentiate between the ground truth and the hallucinated depth images, is:
\begin{multline}
\small
\begin{aligned}
~\mathcal{L}_{D_{depth}} = 
\frac{1}{2} \mathbb{E}_{B_{t} \sim p_{train}(B_{t})} [(D_{depth}(B_{t}) -1)^2] \\
+ \frac{1}{2} \mathbb{E}_{A_{t} \sim p_{train}(A_{t})} [(D_{depth}(G_{A2B}(A_{t}))^2],
\end{aligned}
\label{eq:teach_net_disc}
\end{multline}
where $\mathbb{E}_{B_{t} \sim p_{train}(B_{t})}$ represents a depth image sampled from $p_{train}(B_{t})$, the distribution of depth images in the teacher dataset.

The additional pixel loss, $\mathcal{L}_{pixel}$, between the hallucinated and ground truth depth can be formulated as:
\begin{equation}
\small
\begin{aligned}
\mathcal{L}_{pixel} = \frac{1}{n}\sum_{i=1}^{n}\left | (B_{t})_i - G_{A2B}(A_{t})_i \right |.
\end{aligned}
\label{eq:L_pixel}
\end{equation}
where $n$ is the total number of pixels in an image.

The student component aims to convert a single RGB image $A_{r}$ from the RGB dataset, for which no depth information is available, into a target depth image $\widetilde{B}_{r}$. This is done using the mapping function $G_{A2B}$ from Eq. \ref{eq:teach_net_gen}, along with an inverse mapping function $G_{B2A}$:$B_{r}$$\,\to\,$$A_{r}$, and a discriminator $D_{RGB}$. Loss $\mathcal{L}_{G_{B2A}}$ for the mapping function is formulated as:
\begin{multline}
\small
\begin{aligned}
~\mathcal{L}_{G_{B2A}} & = &
\frac{1}{2} & \mathbb{E}_{A_{r} \sim p_{target}(A_{r})} \\& & &  [(D_{RGB}(G_{B2A}(G_{A2B}(A_{r}))) - 1)^2],
%
\end{aligned}
\label{eq:student_net_gen}
\end{multline}
where $\mathbb{E}_{A_{r} \sim p_{target}(A_{r})}$ represents an RGB image sampled from  $p_{target}(A_{r})$, 
which is the distribution of RGB target dataset.

The loss $\mathcal{L}_{D_{RGB}}$ for the RGB discriminator whose goal is to discriminate between the ground truth RGB $A_r$ and the generated RGB $\tilde{A}_r = G_{B2A}(G_{A2B}(A_r))$, is:
\begin{multline}
\small
\begin{aligned}
~\mathcal{L}_{D_{RGB}} = 
\frac{1}{2} \mathbb{E}_{A_{r} \sim p_{target}(A_{r})} [(D_{RGB}(A_{r}) -1)^2] \\
+ \frac{1}{2} \mathbb{E}_{A_{r} \sim p_{target}(A_{r})} [D_{RGB}(G_{B2A}(G_{A2B}(A_{r})))^2]. 
\end{aligned}
\label{eq:student_net_disc}
\end{multline}

In addition to the supervisory signal from the discriminator, as discussed, we also employ another generator, $G_{B2A}$, to invert the mapping from the hallucinated depth back to RGB. This is done to preserve the identity of the subject and provide additional supervision in a cyclic-consistent way. Accordingly, we formulate the cyclic consistency loss as:
\begin{equation}
\small
\begin{aligned}
\mathcal{L}_{cyc} = \frac{1}{n}\sum_{i=1}^{n}\left | (A_{r})_i - G_{A2B}(G_{B2A}(A_{r}))_i \right |,
\end{aligned}
\label{eq:L_cyc}
\end{equation}
\noindent The total loss for the teacher is then summarized as:
\begin{equation}
\small
\begin{aligned}
\mathcal{L}_{teach} = 
 \mathcal{L}_{G_{A2B}} + \lambda_{pixel} \cdot \mathcal{L}_{pixel},
\end{aligned}
\label{eq:teach_net_tot}
\end{equation}
where $\lambda_{pixel}$ is the weighting parameter for the pixel loss, $\mathcal{L}_{pixel}$, described in Eq. \ref{eq:L_pixel}.

Similarly, the total loss for the student component is summarized as:
\begin{equation}
\small
\begin{aligned}
\mathcal{L}_{student} = 
 \mathcal{L}_{G_{A2B}} + \mathcal{L}_{G_{B2A}} + \lambda_{cyc} \cdot \mathcal{L}_{cyc},
\end{aligned}
\label{eq:student_net_tot}
\end{equation}
where $\lambda_{cyc}$ is the weighting parameter for the cyclic loss, $\mathcal{L}_{cyc}$, described in Eq. \ref{eq:L_cyc}.

The complete training process is listed in pseudocode in Algorithm~\ref{alg:teacher_student}. We first sample an RGB image $A_t$ from $p_{train}(A_t)$ as input to the generator. The output of the generator is the estimated depth image $\widetilde{B_t}$, which is fed to the discriminator and classified as either real or fake. The discriminator is also trained with the corresponding ground truth depth image $B_t$, using the loss mentioned in Eq.~\ref{eq:teach_net_disc}. Apart from the adversarial loss, the training is facilitated with the help of pixel loss (Eq.~\ref{eq:L_pixel}), in the form of MAE loss, for which we define a weighting parameter $\lambda_{pixel}$. 

After training the teacher, we sample an RGB image $A_r$ from the target RGB data $p_{target}(A_r)$, and feed it as input to the generator that is shared between the student and the teacher. The estimated depth images $\widetilde{B_r}$ produced by this generator are then fed to the discriminator in the teacher network stream, thus providing a supervisory signal to generate realistic depth images. These hallucinated depth images are also fed to the inverse generator to transform the estimated depth back into estimated RGB $\widetilde{A_r}$ using the loss mentioned in Eq.~\ref{eq:L_cyc}. As discussed, this is done to preserve the identity information in the depth image while allowing for a more generalized mapping between RGB and depth to be learned through refinement of the original latent RGB-to-D mapping. An additional discriminator, which also follows a fully convolutional structure, is employed to provide an additional supervisory signal for the inverse generator to create realistic RGB images.

\begin{algorithm}[t]
\small
\SetAlgoLined
 \textbf{Input} : teacher dataset $p_{train}(A_{t},B_{t})$, target RGB dataset $p_{target}(A_{r})$, mapping generator function $G_{A2B}$ and $G_{B2A}$, discriminators $D_{RGB}$ and $D_{Depth}$, training configurations (loss weights: $\lambda_{pixel}, \lambda_{cyc}; \text{ learning rates: }\alpha_{teach}, \alpha_{student};$\\ $\text{ decay rate: }\beta_{decay}; \text{ total epochs: } N$);\\
 \While{While n \textless N}{
  Sample $A_{t}, B_{t} \sim p_{train}(A_{t},B_{t})$;\\
  Compute loss $\mathcal{L}_{teach}(A_{t}, B_{t};G_{A2B}, D_{Depth})$ using Eq. \ref{eq:teach_net_tot} and update $G_{A2B}$;\\
  Compute loss $\mathcal{L}_{D_{depth}}(A_{t}, B_{t};G_{A2B})$ using Eq. \ref{eq:teach_net_disc} and update $D_{Depth}$;\\
  Sample $A_{r} \sim p_{target}(A_{r})$;\\
  Compute loss $\mathcal{L}_{student}(A_{r};G_{A2B}, G_{B2A}, D_{RGB})$ using Eq. \ref{eq:student_net_tot} and update $G_{A2B}$ and $G_{B2A}$;\\
  Compute loss $\mathcal{L}_{D_{RGB}}(A_{r};G_{A2B}, G_{B2A})$ using Eq. \ref{eq:student_net_disc} and update $D_{RGB}$;\\
  \textbf{if} $n$ \textgreater epoch teacher \textbf{then}
 $\alpha_{teach} * \beta_{decay}$;\\
  \textbf{else} continue;\\
  \textbf{if} $n$ \textgreater epoch student \textbf{then}
 $\alpha_{student} * \beta_{decay}$;\\
  \textbf{else} continue;
 }
 \caption{Teacher-student learning.}
 \label{alg:teacher_student}
\end{algorithm}

\subsection{Implementation Details}

\textbf{Generator.} We use a fully convolutional structure for the generator inspired by~\cite{zhu2017unpaired}, where an input image of size $128\times128\times3$ is used to output a depth image with the same spatial dimensions.
The encoder part of the generator contains three convolution layers with ReLU activation, where the number of feature maps is gradually increased $(64, 128, 256)$ with a kernel size of $7\times7$ and a stride of $1$ for the first layer. Subsequent layers use a kernel size of $3\times3$ and a stride of $2$. This is followed by $6$ residual blocks, consisting of $2$ convolution layers each with a kernel size of $3\times3$, a stride of $2$, and $256$ feature maps. 
The final decoder part of the generator follows a similar structure, with the exception of using de-convolution layers for upsampling instead of convolution, with decreasing feature maps $(128,64,3)$. The last de-convolution layer which is used to map the features back to images uses a kernel size of $7\times7$ and a stride of $1$, the same as the first layer of the encoder, but with a tanh activation.

\textbf{Discriminator.} We use a fully convolutional architecture for the discriminator, with an input of size $128\times128\times3$. The network uses $4$ convolution layers, where the number of filters gradually increase $(64, 128, 256, 256)$, with a fixed kernel of $4\times4$ and a stride of $2$. All the convolution layers use Instance normalization and leaky ReLU activations with a slope of $0.2$. The final convolution layer uses the same parameters, but with only $1$ feature map.

\textbf{Training.} 
For stabilizing the model, we use the strategy proposed in~\cite{shrivastava2017learning}, updating the discriminators using images from a buffer pool of 50 generated images rather than the ones immediately produced by the generators. Our proposed network is trained from scratch on an Nvidia RTX 2080Ti GPU, using TensorFlow 2.2. We use Adam optimizer and a batch size of 1 as done in~\cite{zhu2017unpaired}. Additionally, we use two different learning rates of $0.0002$ and $0.000002$ for the teacher and student components respectively. Following the suggestions in~\cite{yang2018knowledge}, we start decaying the learning rate for the teacher on the $25^{th}$ epoch with a decay rate $0.5$, sooner than the student, where the learning rate decay starts after the $50^{th}$ epoch. The weights $\lambda_{cyc}$ and $\lambda_{pixel}$ are empirically determined to be $5$ and $10$, respectively.

\section{Experiments}
\subsection{Datasets}
\label{sec:datasets}

\textbf{CurtinFaces}~\cite{mian} is a common RGB-D face dataset which contains over 5000 co-registered RGB and depth image pairs from 52 subjects, captured with a Microsoft Kinect~\cite{zhang2012microsoft}. It has been recorded with varying poses, expressions, and under multiple illumination variations.

\textbf{IIIT-D RGB-D}~\cite{Goswamirgbd,Goswami2014rgbd} contains 4605 RGB-D images from 106 subjects captured using a Microsoft Kinect in two acquisition sessions. Each subject has been captured under normal illumination conditions with variations in pose, expression, and eyeglasses. 
Each image in the dataset is pre-cropped around the face.

\textbf{EURECOM KinectFaceDb}~\cite{min2014kinectfacedb} contains RGB-D face images from 52 people (14 female and 38 male) obtained by a Microsoft Kinect. The data has been captured in 2 different sessions with variations in expression, pose, illumination, and occlusion (a total of 18 images per subject). 

\textbf{Labeled Faces in-the-wild (LFW)}~\cite{huang2008labeled} contains more than 13,000 face images collected from the Internet. Each face has been labeled with the name of the person, with 62 subjects having more than 20 images.

\subsection{Evaluation}
\label{sec:eval}
\textbf{Protocols.} In the training phase, we use the CurtinFaces dataset to train the teacher in order to learn a strict latent mapping between RGB and depth. We choose this dataset as it contains minimal noise among the RGB-D datasets considered in this study, and contains over 5000 co-registered RGB-D images making it the largest. We use its RGB and ground-truth depth images as $A_{t}$ and $B_{t}$ respectively (see Section~\ref{sec:losses}). To train the student, we use the training subsets of the RGB images from IIIT-D RGB-D and EURECOM KinectFaceDb. IIIT-D RGB-D has a pre-defined protocol with a 5-fold cross-validation strategy, to which we strictly adhere. For EURECOM KinectFaceDb, we divide the data into a 50-50 split between the training and testing sets, resulting in a total of 468 images in each set. In the case of the in-the-wild LFW RGB dataset, we utilize 11,953 images for training the generator, and keep the rest of the images for recognition experiments.

For the testing phase of our experiments, we use the trained generator from the student to generate the hallucinated depth images for each RGB image in the test sets. We then further use the RGB and depth images to train the various recognition networks mentioned in Section~\ref{sec:recogntion_results}. For RGB-D datasets, we train the recognition networks on the training sets using the RGB and hallucinated depth images, and evaluate the performance on the test sets. Concerning the LFW dataset, in the testing phase, we use the remaining 20 images from each of the 62 identities that are not used for training. We then use the output RGB and hallucinated depth images as inputs for the recognition experiment.

\textbf{Metrics.}
\label{sec:eval_metric}
We first verify the quality of our depth generation against other generators using pixel-wise quality assessment metrics with respect to the original co-registered ground truth depths. These metrics includes pixel-wise absolute difference, L1 norm, L2 norm, and Root Mean Squared Error (RMSE)~\cite{eigen2014depth,pini2018learning}.
We also use a threshold metric ($\delta$)~\cite{eigen2014depth}, defined as $\% \:\: \text{of} \: y_{i} \: \text{s.t.} \: \max(\frac{y_{i}}{y_{i}^\ast}, \frac{y_{i}^\ast}{y_{i}}) = \delta <  val.$, which measures the percentage of pixels under a certain error threshold, thus providing a similarity score. 
In this metric, $y_{i}$ and $y_{i}^\ast$ represent pixel values in ground truth and hallucinated depths respectively, and $val$ denotes the threshold error value which has been set to $1.25$ as suggested in~\cite{eigen2014depth}. We also use the Fréchet Inception Distance (FID) score~\cite{heusel2017gans} as a measure of similarity between ground truth depth and synthetic depth images.

\textbf{Face Recognition.}
The aim of this study is to use the hallucinated modality to boost recognition performance. As we aim to present results with no dependency on a specific recognition architecture, we use a diverse set of standard deep networks, notably VGG-16~\cite{simonyan2014very}, inception-v2~\cite{ioffe2015batch}, ResNet-50~\cite{he2016deep}, and SE-ResNet-50~\cite{hu2018squeeze} in our evaluation. We report the rank-1 identification results with and without ground truth depth for RGB-D datasets as well as the results obtained by the combination of RGB and our hallucinated depth images. For LFW RGB dataset, we naturally do not have ground truth depths, so we only present the identification results with and without our hallucinated depth. We also use different strategies, including feature-level fusion, score-level fusion, two-level attention fusion~\cite{uppal2020attention}, and depth-guided attention~\cite{uppal2021depth}, when combining RGB and depth images.

\section{Performance}
\subsection{Quality Assessment}
\label{sec:comp_study}

\begin{figure}
    \centering
    \includegraphics[width = .85\columnwidth]{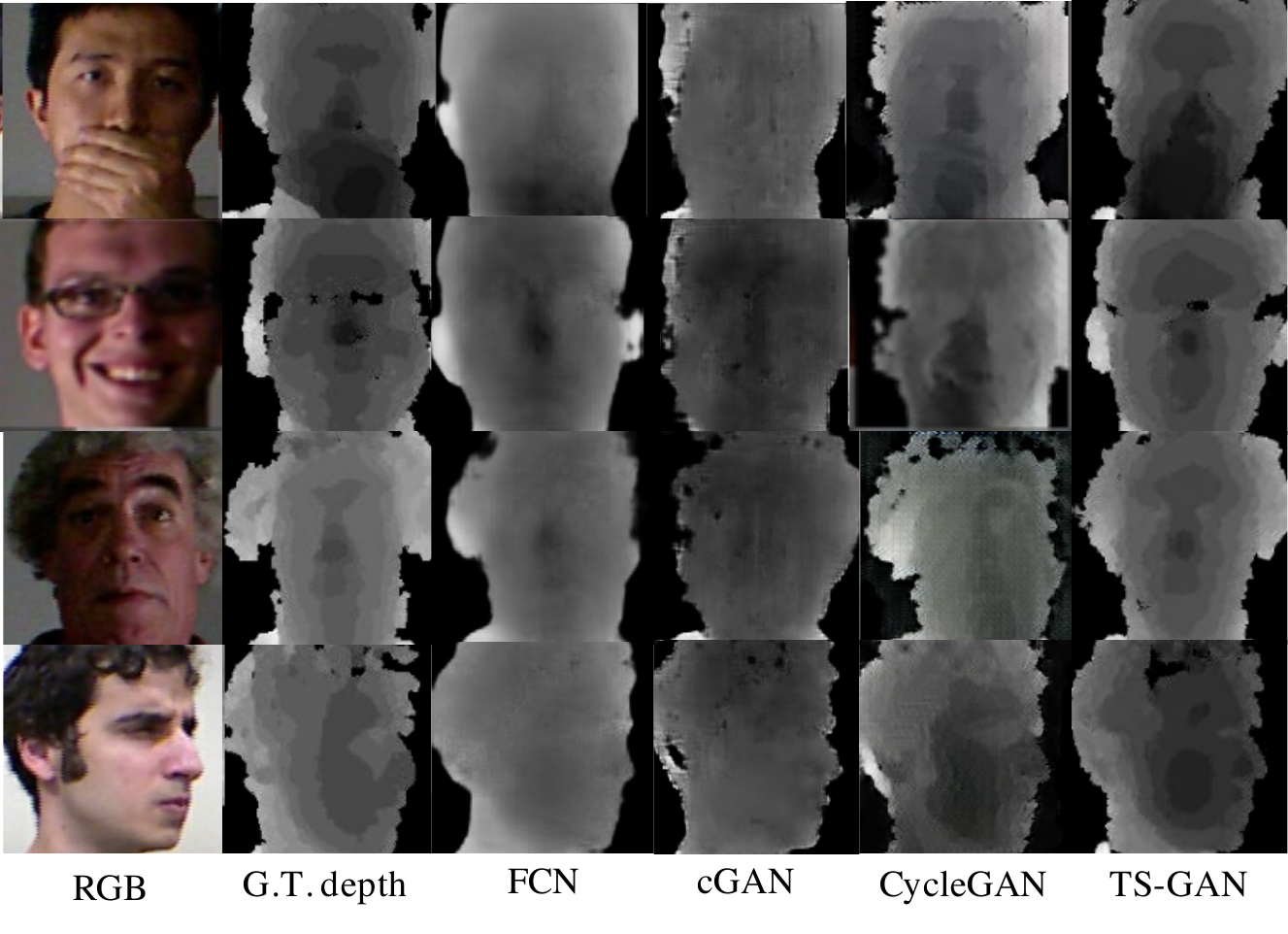}
    \caption{Several input RGB test samples from the CurtinFaces dataset along with ground truth (G.T.) co-registered depth images, and synthesized depth images generated by various state-of-the-art alternatives and our proposed method are presented.}
    \label{fig:depth_comp}
\end{figure}

We first compare the performance of TS-GAN with alternative depth generators, namely Fully Convolutional Network (FCN)~\cite{cui2018improving_newadd}, image-to-image translation cGAN~\cite{pini2018learning}, and CycleGAN~\cite{kwak2020novel}
To this end, we perform experiments on the CurtinFaces dataset, where we use 47 out of the 52 subjects for training the generator, and use the remaining 5 subjects for generating depth images to be used for quality assessment experiments. Figure~\ref{fig:depth_comp} shows depth generated by the alternative methods as well as our TS-GAN on some of the test subjects. As can be seen, our method is able to generate realistic depth images which appear very similar to the ground truth depth images. 

\begin{figure}
    \centering
    \includegraphics[width=0.7\columnwidth]{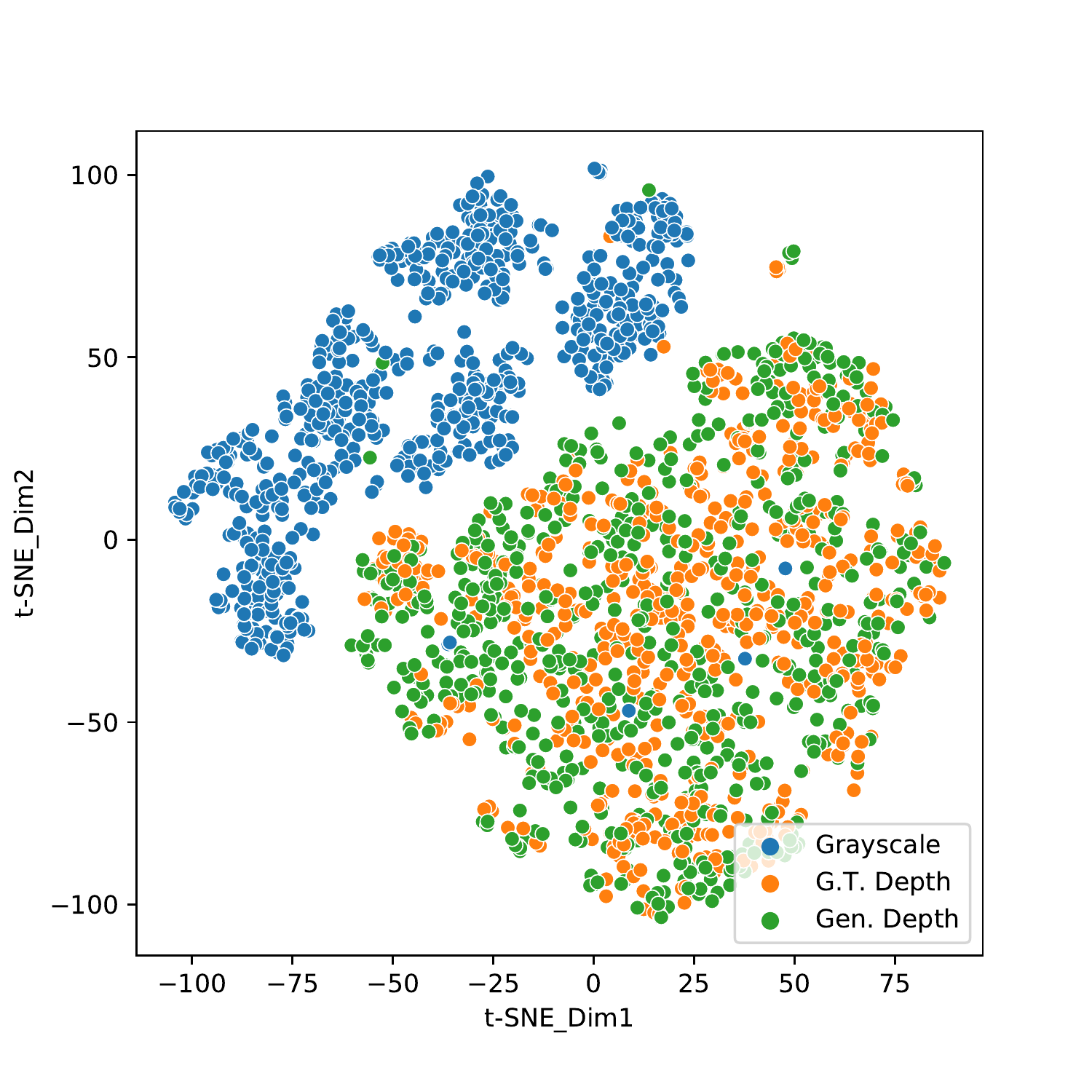}
    \caption{A t-SNE visualization of input RGB images (converted to grayscale), Ground Truth (G.T.) depth, and the output depth Hallucinated (Hal.) by TS-GAN.}
    \label{fig:rgbd_tsne}
\end{figure}

In Figure~\ref{fig:rgbd_tsne} we present a t-SNE~\cite{van2008visualizing} visualization of embeddings generated by a ResNet-50 network for a number of RGB samples from CurtinFaces (converted to grayscale in order for color to not be considered a factor), ground truth depth images, and hallucinated depth images by TS-GAN. This figure demonstrates a very high degree of overlap between the ground truth and generated depth images, thus depicting their similarity.

Table \ref{tab:comp_methods} shows the results for pixel-wise objective metrics (Section \ref{sec:eval_metric}). For the first four metrics namely absolute difference, L1 Norm, L2 Norm, and RMSE, lower values indicate better image quality. It can be observed that our proposed method mostly outperforms the other methods, the single exception being the absolute difference metric in which FCN shows slightly better performance. A potential reason for this anomaly is that FCN only uses one loss function that aims to minimize the absolute error between the ground truth and the generated depth, naturally resulting in minimal absolute difference error. For the threshold metric $\delta$, the higher percentage of pixels under the threshold error value of 1.25 achieved by our method represents better spatial accuracy for the generated depth images. 
Lastly, the lower obtained FID scores indicate that the proposed method images are most similar to the ground truth depth samples.

\begin{table}[]
    \centering
    \scriptsize
    \setlength\tabcolsep{5pt}
    \caption{Comparisons of image quality metrics between our method and other depth generation methods.}
    \begin{tabular}{|l|c|c|c|c|}
    \hline
         \textbf{Metrics} & \textbf{FCN} \cite{cui2018improving_newadd} &  \textbf{cGAN \cite{pini2018learning}} & \textbf{CycleGAN} \cite{kwak2020novel} & \textbf{Ours (TS-GAN)} \\
         \hline\hline
         Abs. Diff. $\downarrow$  & \textbf{0.0712} & 0.0903 & 0.1037 & 0.0754 \\
         L1 Norm $\downarrow$ & 0.2248 & 0.2201 & 0.2387 & \textbf{0.2050}\\
         L2 Norm $\downarrow$ & 89.12 & 89.05 & 90.32 & \textbf{82.54} \\
         RMSE $ \downarrow$ & 0.3475 & 0.3474 & 0.3542 & \textbf{0.3234}\\
         $\delta(1.25)  \uparrow$ & 64.31 & 64.27 & 65.76 & \textbf{69.02} \\
         $\delta(1.25^{2})  \uparrow $& 81.66 & 82.08 & 82.56 & \textbf{87.20}\\
         $\delta(1.25^{3})  \uparrow $& 94.33 & 95.10 & 95.63 & \textbf{97.54}\\
         FID $\downarrow$  & 17.72 & 16.39 & 16.13 & \textbf{14.67} \\
         \hline
    \end{tabular}
    \label{tab:comp_methods}
\end{table}

\begin{figure}
    \centering
    \includegraphics[width=\columnwidth]{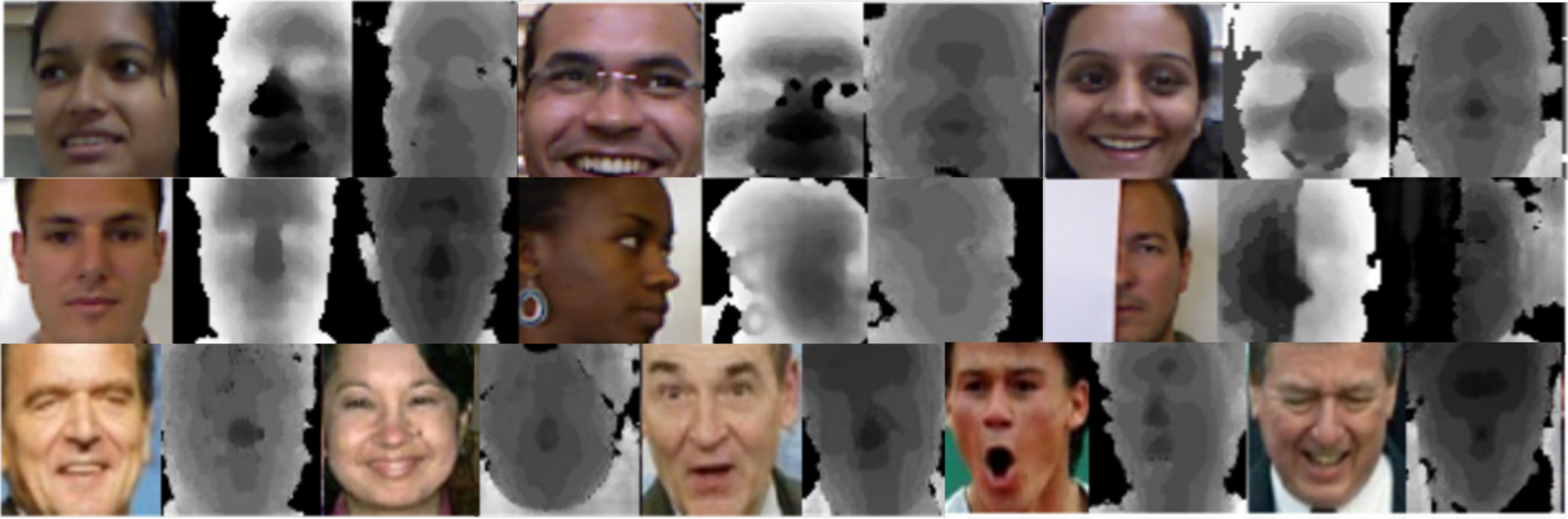}
    \caption{The first two rows show samples of the RGB-D datasets (IIIT-D and EURECOM KinectFaceDb). The first column shows RGB images, the second column shows the ground truth depth, and the third column shows the hallucinated depth. In the third row LFW samples are presented where the first column shows the RGB images while the second column shows the hallucinated depth.}
    \label{fig:3ds_results}
\end{figure}

In order to show the generalization of our generator when applied to the target datasets (mentioned in Section ~\ref{sec:method}) for testing, hallucinated depth samples for IIIT-D and EURECOM RGB-D datasets are shown in Figure~\ref{fig:3ds_results} (top and middle rows). The first and second columns show the input RGB images and the ground truth depth image corresponding to the RGB image, while the third column shows the generated depth images. As can be seen, our methods can adopt to different poses, expressions and occlusions present in the target datasets. The bottom row in this figure shows the depth generated for the in-the-wild LFW RGB dataset, where our method is able to adopt to the non-frontal and unnatural poses which are not present in the constrained, lab-acquired RGB-D datasets.

\begin{table}[]
    \centering
    \scriptsize
    \setlength
    \tabcolsep{3.3pt}
        \caption{IIIT-D and EURECOM KinectFaceDb rank-1 recognition results. $\widetilde{\text{D}}$ denotes the hallucinated depth using TS-GAN. 
        }
    \begin{tabular}{|l|l|c|c|c|c|c|}
    \hline
        & &\multicolumn{5}{|c|}{\textbf{Accuracy}}\\\cline{3-7}
         \textbf{Dataset} & \textbf{Model} &  \textbf{RGB} & \multicolumn{2}{|c|}{\textbf{RGB + D}}&\multicolumn{2}{|c|}{\textbf{RGB + $\Dtilde$}}\\ 
         \cline{3-7}
         & &  \multirow{2}{*}{\textbf{--}} & \textbf{Feat.}& \textbf{Score} & \textbf{Feat.}&\textbf{Score} \\
          & & & \textbf{Fusion}& \textbf{Fusion} & \textbf{Fusion}&\textbf{Fusion} \\
         \hline\hline
         \multirow{6}{*}{IIIT-D} & VGG-16 & 94.1\% & \textbf{95.4}\% & 94.4\% & \textbf{95.4}\% & 94.1\% \\
         & Inception-v2 & 95.0\% &\textbf{96.5}\% & 95.0\% & 96.1\% & 95.9\% \\
         
         & ResNet-50 & 95.8\% & 96.9\%& 95.9\% &\textbf{97.1}\% & 96.1\% \\
         
         & SE-ResNet-50 & 97.8\% & \textbf{98.9}\% & 97.9\% & 98.6\% & 97.6\% \\
         
        & Two-level att.~\cite{uppal2020attention} & -- & \textbf{99.4}\% & -- & 99.1\% & -- \\
         
         & Depth-guid. att.~\cite{uppal2021depth} & -- & \textbf{99.7}\% & -- & \textbf{99.7}\% & -- \\
         \hline
         \multirow{6}{*}{EURECOM} & VGG-16 & 83.6\% & \textbf{88.4}\% & 84.5\% & 88.3\% & 84.2\% \\
         & Inception-v2 & 87.5\% & \textbf{90.3}\% & 86.9\% & 90.1\% & 87.9\% \\
         & ResNet-50 & 90.8\% & 92.1\% & 91.0\% & \textbf{92.2}\% & 90.7\%\\
         & SE-ResNet-50 & 91.3\% & 93.1\% & 91.6\% & \textbf{93.2}\% & 91.5\%\\
         & Two-level att.~\cite{uppal2020attention} & -- & 92.0\% & -- & \textbf{92.3}\% & --\\
         & Depth-guid. att.~\cite{uppal2021depth} & -- & \textbf{93.1}\% & -- & 92.7\% & --\\
         \hline 
    \end{tabular}
    \vspace{2pt}
    \label{tab:IIITD_Eurecom_fr_results}
\end{table}

\begin{table}
    \centering
    \scriptsize
    \setlength
    \tabcolsep{5pt}
        \caption{EURECOM KinectFaceDb pose and occlusion test set recognition.}
    \begin{tabular}{|l|l|c|c|c|}
    \hline
        & &\multicolumn{3}{c|}{\textbf{Accuracy}}\\\cline{3-5}
         \textbf{Test set} & \textbf{Model} &  \textbf{RGB} & \multicolumn{1}{c|}{\textbf{RGB + D}}&\multicolumn{1}{c|}{\textbf{RGB + $\Dtilde$}}\\ 
         \cline{3-5}
         \hline\hline
         \multirow{6}{*}{Left Pose Set} & VGG-16 & 75.2\% & \textbf{77.4}\% &  77.2\%  \\
         & Inception-v2 & 75.8\% & \textbf{78.1}\% & 77.6\%  \\
         & ResNet-50 & 77.4\% & 80.4\%  & \textbf{80.5}\% \\
         & SE-ResNet-50 & 79.2\% & 80.8\%  & \textbf{81.1}\% \\
         & Two-level att. & -- & \textbf{81.6}\% & 81.3\% \\
         & Depth-guid. att. & -- & 82.5\%  & \textbf{82.7}\% \\
         \hline 
          \multirow{6}{*}{Right Pose Set} & VGG-16 & 74.8\% & \textbf{77.6}\% &  77.5\%  \\
         & Inception-v2 & 75.9\% & \textbf{78.6}\% & 78.4\%  \\
         & ResNet-50 & 77.2\% & 80.1\%  & \textbf{80.3}\% \\
         & SE-ResNet-50 & 78.9\% & 80.4\%  & \textbf{80.7}\% \\
         & Two-level att. & -- & \textbf{81.9}\% & 81.5\% \\
         & Depth-guid. att. & -- & \textbf{82.6\%}  & 82.3\% \\
         \hline 
         \multirow{6}{*}{Occlusion Set} & VGG-16 & 84.8\% & \textbf{87.4}\% & 87.2\%  \\
         & Inception-v2 & 86.2\% & 88.3\%  & \textbf{89.8}\%  \\
         & ResNet-50 & 88.9\% & 90.1\%  & \textbf{90.8}\%\\
         & SE-ResNet-50 & 90.8\% & 92.2\%  & \textbf{92.5}\%\\
         & Two-level att. & -- & \textbf{92.5}\%  & \textbf{92.5}\% \\
         & Depth-guid. att. & -- & \textbf{93.8}\%  & 93.2\% \\
         \hline 
    \end{tabular}
    \vspace{2pt}
    \label{tab:pose_results}
\end{table}

\subsection{Recognition Results}
\label{sec:recogntion_results}

\textbf{RGB-D Datasets.} To demonstrate the effectiveness of the hallucinated depth for face recognition, the mapping function (Eq.~\ref{eq:teach_net_gen}) is used to estimate the corresponding depth images for the RGB images,
both of which are used as input to the recognition pipeline. Table~\ref{tab:IIITD_Eurecom_fr_results} shows the rank-1 recognition results on the IIIT-D and KinectFaceDb datasets using the four networks discussed earlier. We have considered different fusion strategies as well as two recent attention-based RGB-D solutions~\cite{uppal2020attention,uppal2021depth} as mentioned in Section \ref{sec:eval}. It can be observed that the fusion of RGB and the depth hallucinated using TS-GAN constantly provides better results across all the CNN architectures, when compared to using only the RGB images.

For further comparison, we also perform recognition with RGB and the ground truth depth using the same pipelines. For the IIIT-D dataset, recognition with RGB and generated depth leads to comparable results to that with RGB and ground truth depth images. Concerning the EURECOM KinectFaceDb dataset, the results also show that our generated depth provide added value to the recognition pipeline as competitive results (slightly below) to that of RGB and ground truth depth are achieved. Interestingly, in some cases for both IIIT-D and KinectFaceDb, our hallucinated depth images even provide superior performance over the ground-truth depth images. This is most likely due to the fact that some depth images available in the IIIT-D and KinectFaceDb datasets are noisy, while our generator can provide cleaner synthetic depth images as it has been trained on higher quality depth images available in the CurtinFaces dataset.  
Finally, to test the robustness of our approach on variations in pose and occlusions, we perform experiments using the EURECOM KinectFaceDb dataset. The results presented in Table~\ref{tab:pose_results} indicate that our TS-GAN results in high quality depth images even with variations in pose and occlusions, as evidenced by the high recognition rates.

\textbf{RGB Dataset.} 
Table~\ref{tab:LFW_fr_results} presents the recognition results on the in-the-wild LFW dataset, where the results are presented both with and without our hallucinated depth images. We observe that the hallucinated depth significantly improves the recognition accuracy across all the CNN architectures, with $3.4\%$, $2.4\%$, $2.3\%$, $2.4\%$ improvements for VGG-16, Inception-v2, ResNet-50, and SE-ResNet-50 respectively. The improvements are more obvious when considering the state-of-the-art attention-based methods, clearly demonstrating the benefits of our synthetic depth images to improve recognition accuracy.

\begin{table}[]
    \centering
    \scriptsize
    \setlength\tabcolsep{5pt}
        \caption{LFW rank-1 recognition results. $\widetilde{\text{D}}$ denotes the hallucinated depth using TS-GAN. 
        }
    \begin{tabular}{|l|c|c|c|}
    \hline
        & \multicolumn{3}{|c|}{\textbf{Accuracy}}\\\cline{2-4}
         \textbf{Model} & \textbf{RGB} & \multicolumn{2}{|c|}{\textbf{RGB + $\Dtilde$}}\\ 
         \cline{2-4}
           & \multirow{2}{*}{\textbf{--}} & \textbf{Feature}&\textbf{Score} \\
           &  & \textbf{Fusion}&\textbf{Fusion} \\
         \hline\hline
         VGG-16 & 75.3\% & \textbf{78.7}\% & 76.1\% \\
         Inception-v2 & 78.1\% & \textbf{80.5}\% & 78.4\% \\
         ResNet-50 & 81.8\% & \textbf{84.1}\% & 81.7\% \\
         SE-ResNet-50 & 83.2\% & \textbf{85.6}\% & 83.2\% \\
         Two-level att.~\cite{uppal2020attention} & -- & \textbf{84.7}\% & -- \\
         Depth-guided att.~\cite{uppal2021depth} & -- & \textbf{85.9}\% & -- \\
         \hline
    \end{tabular}
    \vspace{2pt}
    \label{tab:LFW_fr_results}
\end{table}

\begin{table}[]
    \centering
    \scriptsize
    \caption{Ablation study on IIIT-D and EURECOM KinectFaceDb. }
    \begin{tabular}{|l|l|l|c|c|}
    \hline
         & & \textbf{IIIT-D} & \textbf{KinectFaceDb}\\
         \textbf{Ablation Model} &\textbf{Classifier} & \textbf{Accuracy}& \textbf{Accuracy}\\
         \hline\hline
        \multirow{4}{*}{Teacher} & \multirow{1}{*}{VGG-16} & \textbf{95.4}\% & 85.7\%\\
         & \multirow{1}{*}{Inception-v2} & 95.0\% & 88.6\%\\
         & \multirow{1}{*}{ResNet-50} & 96.6\% & 91.3\%\\
         & \multirow{1}{*}{SE-ResNet-50} & 98.4\% & 91.9\%\\
         \hline
        \multirow{4}{*}{Teacher's A2B Gen.} & \multirow{1}{*}{VGG-16}& 95.1\% & 87.8\%\\
         & \multirow{1}{*}{Inception-v2} & 96.0\% & 88.2\%\\
         & \multirow{1}{*}{ResNet-50} & 96.7\% & 90.6\%\\
         & \multirow{1}{*}{SE-ResNet-50} & 98.5\% & 92.2\%\\
         \hline
         \multirow{4}{*}{\textbf{Teacher-Student (TS-GAN)}} & \multirow{1}{*}{VGG-16} &\textbf{95.4}\% & \textbf{88.3}\%\\
         & \multirow{1}{*}{Inception-v2} & \textbf{96.1}\% & \textbf{90.1}\%\\
         & \multirow{1}{*}{ResNet-50} & \textbf{97.1}\% & \textbf{92.2}\%\\
         & \multirow{1}{*}{SE-ResNet-50} & \textbf{98.6}\% & \textbf{93.2}\%\\
         \hline
    \end{tabular}
    
    \label{tab:ablation}
\end{table}

\subsection{Ablation study}
To evaluate the impact of each of the main components of our solution, we perform ablation experiments by systematically removing them. First, we remove the student component, leaving just the teacher. Next, we remove the discriminator from the teacher leaving only the A2B generator as discussed in Section \ref{sec:method} (also see Figure \ref{fig:architecture}). The results are presented in Table~\ref{tab:ablation} and compared to our complete TS-GAN solution. The presented recognition results are obtained using a feature-level fusion scheme to combine RGB and hallucinated depth images. The results show that performance suffers by the removal of each component for all four CNN architectures, demonstrating the effectiveness of our approach.

\begin{figure}
    \centering
    \includegraphics[width=.7\columnwidth]{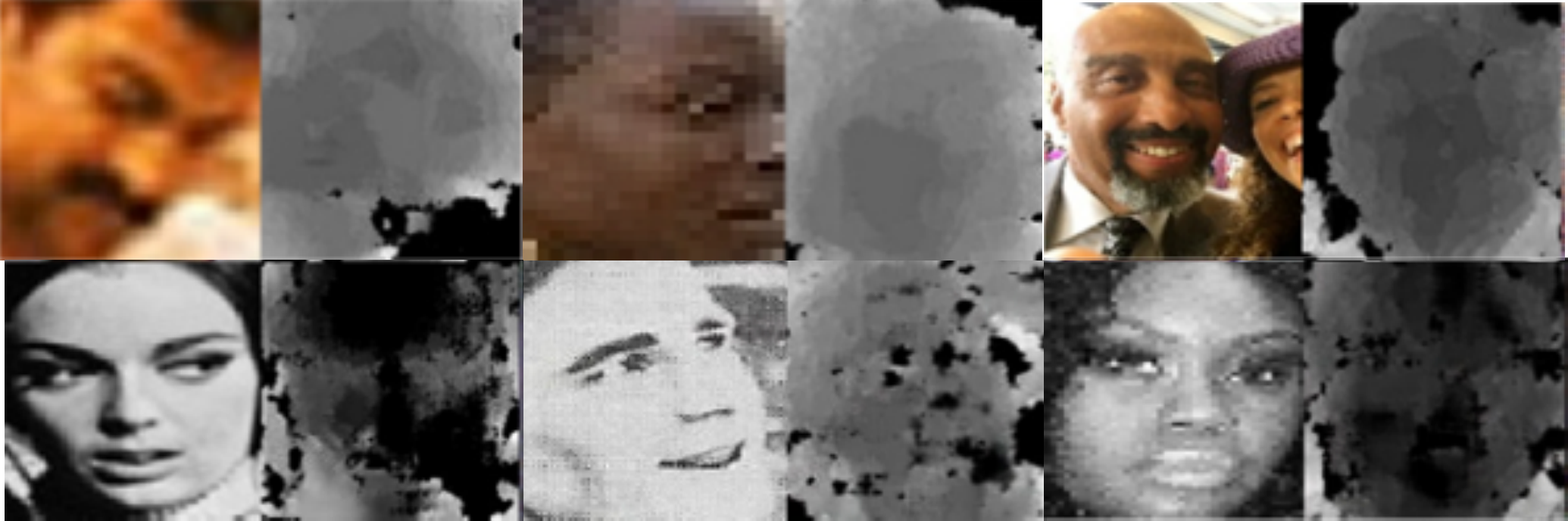}
    \caption{A few samples of failed depth hallucination.}
    \label{fig:failure_cases}
\end{figure}

\section{Limitations}

Although our proposed method provides impressive results, high quality depth images can not be generated in some cases. 
Our proposed method has been successful in adopting to various non-frontal poses and expressions as can be seen in Figure \ref{fig:3ds_results}, however, it maintains some sensitivity to the diversity of the training data. For instance, gray-scale images were not used to train the generator, and hence adopting to them proves difficult for the generator as seen in Figure \ref{fig:failure_cases}. Our method also fails to generate high quality depth images for very low-resolution images and multiple faces in the same image. To mitigate these problems, we could create a larger and more diverse training set to include these variations during training which could help the generator with better generalization.

\section{Conclusion}
In this paper, we propose a novel teacher-student adversarial architecture for depth generation from RGB images, called TS-GAN, to boost the performance of facial recognition systems. The teacher component of our method consisting of a generator and a discriminator learns a strict latent mapping between RGB and depth image pairs following a supervised approach. The student, which itself consists of a generator-discriminator pair along with the generator shared with the teacher, then refines this mapping by learning a more generalized relationship between the RGB and depth domains for samples without corresponding co-registered depth images. Comprehensive experiments on three public face datasets show that our method outperformed other depth generation methods.

{\small
\noindent \textbf{Acknowledgements.} The authors would like to thank Irdeto Canada Corporation and the Natural Sciences and Engineering Research Council of Canada (NSERC) for funding this research.}

{\small
\bibliographystyle{ieee_fullname}
\bibliography{egbib}
}

\end{document}